\documentclass{article}

\usepackage[square,numbers]{natbib}

\usepackage[final]{nips_2017}

\usepackage[utf8]{inputenc} % allow utf-8 input
\usepackage[T1]{fontenc}    % use 8-bit T1 fonts
\usepackage{hyperref}       % hyperlinks
\usepackage{url}            % simple URL typesetting
\usepackage{booktabs}       % professional-quality tables
\usepackage{amsfonts}       % blackboard math symbols
\usepackage{nicefrac}       % compact symbols for 1/2, etc.
\usepackage{microtype}      % microtypography
\usepackage{amsmath}
\usepackage[pdftex]{graphicx} 
\graphicspath{{figures/}}
\usepackage{float}
\floatstyle{plaintop}
\restylefloat{table}

\usepackage{algorithm}
\usepackage{algorithmic}
\usepackage{amssymb}

\usepackage[tableposition = top]{caption}

\title{Improved Neural Text Attribute Transfer with Non-parallel Data}

\author{
  Igor Melnyk\thanks{Corresponding author. Email: \texttt{igor.melnyk@ibm.com}} \And Cicero Nogueira dos Santos \And Kahini Wadhawan \And Inkit Padhi \And Abhishek Kumar\\
  \\
  IBM Research AI\\
  T. J. Watson Research Center\\
  Yorktown Heights, NY
  \vspace{-10pt}
}

\begin{document}
\maketitle

\begin{abstract}
Text attribute transfer using non-parallel data requires methods that can perform disentanglement of content and linguistic attributes. 
In this work, we propose multiple improvements over the existing approaches that enable the encoder-decoder framework to cope with the text attribute transfer from non-parallel data. We perform experiments on the sentiment transfer task using two datasets. For both datasets, our proposed method outperforms a strong baseline in two of the three employed evaluation metrics.
% Text style transfer delves into changing linguistic attributes of the text while making sure that the content is retained. This work extends the behavior of Natural Language Generation(NLG) systems to incorporate the desired style while generating the text and yet preserving the content. Our method builds upon the encoder decoder architecture that disentangles style from the content of the text. However, as we don't usually have parallel data, we propose a novel method of using collaborative classifier and various content preservation losses to solve the problem. We analyze the effectiveness of our approach using sentiment transfer tasks which uses non-parallel monolingual data.
\end{abstract}

\section{Introduction}
\label{sec:introduction}
%\vspace{-5pt}
The goal of the \emph{text attribute transfer} task is to change an input text such that the value of a particular linguistic attribute of interest (e.g. language = English, sentiment = Positive) is transferred to a different desired value (e.g. language = French, sentiment = Negative).
This task needs approaches that can disentangle the content from other linguistic attributes of the text.
The success of neural encoder-decoder methods to perform text attribute transfer for the tasks of machine translation and text summarization rely on the use of large parallel datasets that are expensive to be produced.
The effective use of non-parallel data to perform this family of problems is still an open problem.

In text attribute transfer from non-parallel data,
given two large sets of non-parallel texts $X_0$ and $X_1$, which contain different attribute values $s_0$ and $s_1$, respectively, the task consists in using the data to train models that can rewrite a text from $X_0$ such that the resulting text has attribute value $s_1$, and vice-versa. The overall message contained in the rewritten text must be relatively the same of the original one, only the chosen attribute value should change.
Two of the main challenges when using non-parallel data to perform such task are: (a) 
there is no  straightforward way to train the encoder-decoder because we can not use maximum likelihood estimation on the transferred text due to lack of ground truth;
(b) it is difficult to preserve content while transferring the input to the new style. 
Recent work from \citet{shen17} showed promising results on style-transfer from non-parallel text by tackling challenging (a).

In this work, we propose a new method to perform text attribute transfer that tackles both challenges (a) and (b).
We cope with (a) by using a single collaborative classifier, as an alternative to commonly used adversarial discriminators, e.g., as in \cite{shen17}. Note that a potential extension to a problem of multiple attributes transfer would still use a single classifier, while in \cite{shen17} this may require as many discriminators as the number of attributes.
We approach (b) with a set of constraints, including the attention mechanism combined with cyclical loss and a novel noun preservation loss to ensure proper content transfer. We compared our algorithm with \citet{shen17} on the sentiment transfer task on two datasets using three evaluation metrics (sentiment transfer accuracy, a novel \emph{content preservation} metric and a perplexity), outperforming the baseline in terms of the first two.

\section{Proposed Method}
\label{sec:method}
\begin{figure}[!t]
	\centering
	\includegraphics[width=\textwidth]{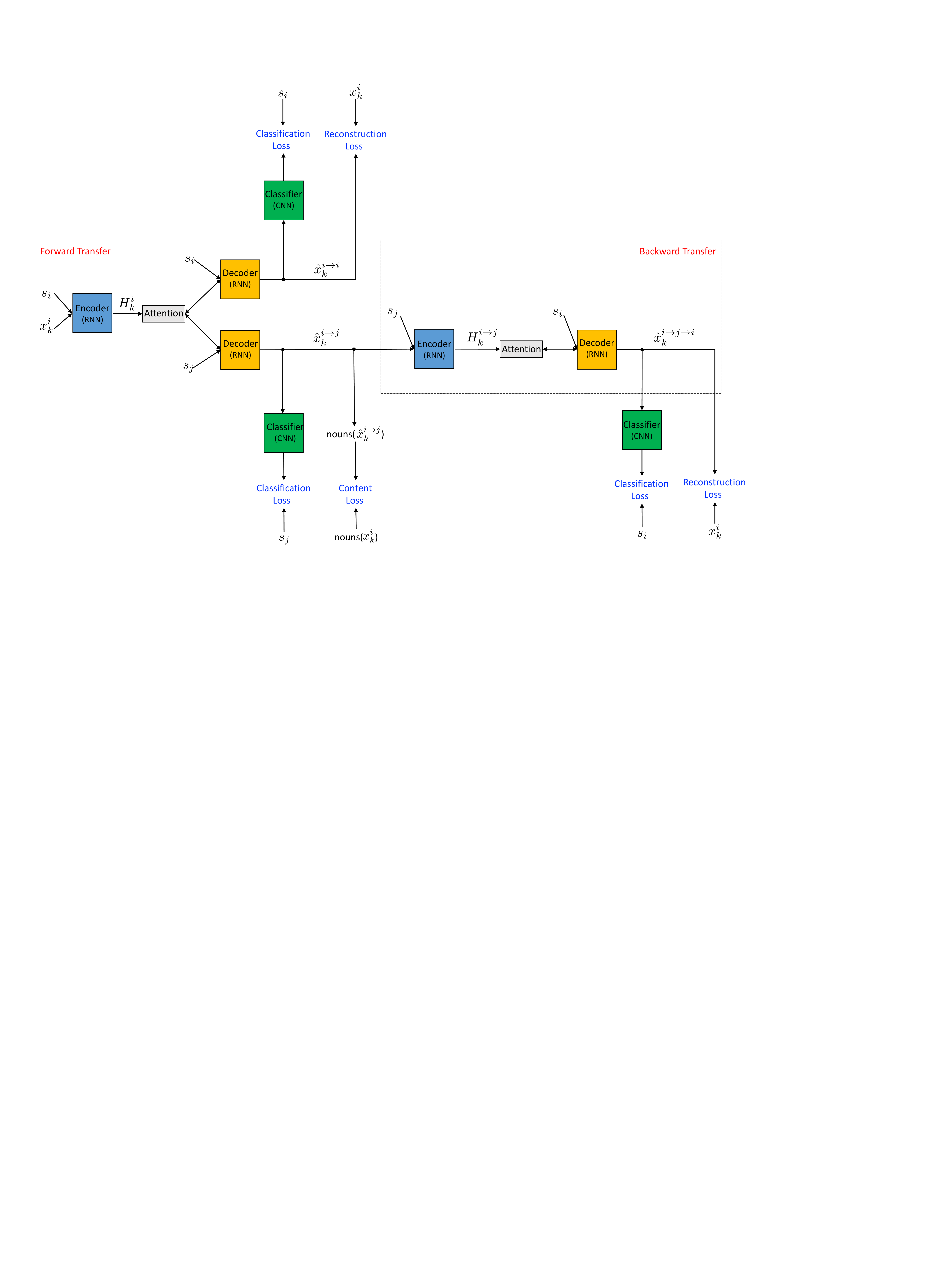}
	\caption{Proposed framework of a Neural Text Attribute Transfer algorithm using non-parallel data.}
	\label{fig:model_training}
\end{figure}

We assume access to a text dataset consisting of two non-parallel corpora $X = X_0 \cup X_1$ with different attribute values $s_0$ and $s_1$ of a total of $N=m+n$ sentences, where $|X_0| = m$ and $|X_1|=n$. We denote a randomly sampled sentence $k$ of attribute $s_i$ from $X$ as $x_k^{i}$, for $k \in 1, \ldots, N$ and $i \in \{0,1\}$. 
A natural approach to perform text attribute transfer is to use a regular encoder-decoder network, however, the training of such network requires parallel data. Since in this work we consider a problem of attribute transfer on non-parallel data, we propose to extend the basic encoder-decoder by introducing a collaborative classifier and a set of specialized loss functions that enable the training on such data. Figure \ref{fig:model_training} shows an overview of the proposed attribute transfer approach. Note that for clarity in the Figure \ref{fig:model_training} we have used multiple boxes to show encoder, decoder and classifier, the actual model contains a single encoder and decoder, and one classifier.

The encoder (in the form of RNN), $E(x_k^{i}, s_i)=H_k^i$, takes as input a sentence $x_k^{i}$ together with its attribute label $s_i$, and outputs $H_k^i$, a sequence of hidden states.
%, whose length is the same as the length of the input sentence $x_k^{i}$.
%We implement the encoder using a recurrent neural network (RNN) that uses the style label as its initial state.
The decoder/generator (also in the form of RNN), $G(H_k^i, s_j) = \hat{x}_k^{i\rightarrow j}$ for $i,j \in {0,1}$, takes as input the previously computed $H_k^i$ and a desired attribute label $s_j$ and outputs a sentence $\hat{x}_k^{i\rightarrow j}$, 
which is the original sentence but transferred from attribute value $i$ to attribute value $j$. The hidden states $H_k^i$ are used by the decoder in the attention mechanism \cite{luong15,bahdanau14}, and in general can improve the quality of the decoded sentence. For $i=j$, the decoded sentence $\hat{x}_k^{i\rightarrow i}$ is in its original attribute $s_i$ (top part of Figure \ref{fig:model_training}); for $i \neq j$, the decoded/transferred sentence $\hat{x}_k^{i\rightarrow j}$ is in a different attribute $s_j$ (bottom part of Figure\ref{fig:model_training}). Denote all transferred sentences as $\hat{X} = \{\hat{x}_k^{i\rightarrow j} ~|~ i\neq j, k=1,\ldots, N\}$. The classifier (in the form of CNN), then takes as input the decoded sentences and outputs a probability distribution over the attribute labels, i.e., $C(\hat{x}_k^{i\rightarrow j}) = p_C(s_j|\hat{x}_k^{i\rightarrow j})$ (see Eq. \eqref{eq:class_td} for more details). By using the collaborative classifier our goal is to produce a training signal that indicates the effectiveness of the current decoder on transferring a sentence to a given attribute value.

%When all the sentences in $X$ are transferred to the opposite style, we denote the result by $$. 
%The decoder is also implemented in the form of a RNN, whose initial state is defined by style label $s_j$. 

Note that the top branch of Figure \ref{fig:model_training} can be considered as an auto-encoder and therefore we can enforce the closeness between $\hat{x}_k^{i\rightarrow i}$ and $x_k^i$ by using a standard cross-entropy loss (see \eqref{eq:rec_loss} below). However, for the bottom branch, due to lack of parallel data, we cannot use the same approach, and for this purpose we proposed a novel content preservation loss (see Eq. \eqref{eq:cnt_loss}). Finally, note that once we transferred $X$ to $\hat{X}$ (forward-transfer step), we can now transfer $\hat{X}$ back to $X$ (back-transfer step) by using the bottom branch in Figure \ref{fig:model_training} (see Eq. \eqref{eq:back_rec_loss} and Eq. \eqref{eq:class_btd} below). 

%The outputs of the forward step are used to compute different loss functions to train both the encode-decoder and the classifier, which are trained together in an end-to-end fashion.

%\begin{figure}[!ht]
%	\centering
%	\includegraphics[width=0.7\textwidth]{Figures/tst_forward_step_single}
%	\caption{.}
%	\label{fig:model_test}
%\end{figure}
In what follows, we present the details of the loss functions employed in training of our model.

%\begin{algorithm}[!th]
%\caption{Training of the Neural Style Transfer algorithm on non-parallel data.}
%\label{alg:train}
%\begin{algorithmic}
%\REQUIRE Two non-parallel corpora $X = X_0 \cup X_1$ with different styles $s_0$ and $s_1$.
%\STATE Initialize $\theta_E,\theta_G,\theta_{C}$
%\REPEAT

%\STATE Sample a mini-batch of $l$ examples $A=\{x_k^{i}\}_{k=1}^l$ from $X$, with $i \in \{0,1\}$
%\STATE Sample a mini-batch of $l$ examples $B=\{\hat{ x}_k^{i\rightarrow j}\}_{k=1}^l$ from the generator's distribution $p_G$, where $\hat{x}_k^{i\rightarrow j} = G(E( x_k^i,s_i),s_j)$ with $i,j \in \{0,1\}$

%\STATE Compute the reconstruction loss $\mathcal L_{\text{rec}}$ by Eq. (\ref{eq:rec_loss})
%\STATE Compute back-transfer loss $\mathcal L_{\text{back\_rec}}$ by Eq. (\ref{eq:back_rec_loss})
%\STATE Compute classification losses $\mathcal L_{class\_od}$, $\mathcal L_{class\_td}$ and $\mathcal L_{class\_btd}$ by Eqs. (\ref{eq:class_od}), (\ref{eq:class_td}) and (\ref{eq:class_btd}), respectively.

%\STATE Update $\{\pmb\theta_E,\pmb\theta_G\}$ by gradient descent on loss $\mathcal L_{rec} + \mathcal L_{back\_rec} + \mathcal L_{class\_td} + \mathcal L_{class\_btd}$
%\STATE Update $\pmb\theta_{C}$ by gradient descent on loss $\mathcal L_{class\_od} + \mathcal L_{class\_td} + \mathcal L_{class\_btd}$
%\UNTIL{convergence}
%\end{algorithmic}
%\end{algorithm}

\subsection{Forward Transfer}
\textbf{Reconstruction Loss.} 
%the word \emph{sequence reconstruction loss} is the usual negative log-likelihood loss used in sequence-to-sequence models.
%In our case we can also call it the auto-encoder loss since it is exactly the type of loss normally used in sequence-to-sequence auto-encoders.
Given the encoded input sentence $x_k^{i}$ and the decoded sentence $\hat{x}_k^{i\rightarrow i}$, the reconstruction loss measures how well the decoder $G$ is able to reconstruct it:
\begin{align}
\label{eq:rec_loss}
\mathcal L_{rec} = \mathbb{E}_{x_k^i\sim X} \left[-\log p_G(x_k^i| E(x_k^i, s_i), s_i)\right].
\end{align}

\textbf{Content Preservation Loss.}
To enforce closeness between $x_k^{i}$ and $\hat{x}_k^{i\rightarrow j}$ for $i\neq j$, we utilize the attention mechanism. Recall, that this mechanism enables to establish an approximate correspondence between the words in the original (encoded) and transferred (decoded) sentences. For example, denote the words in sentence $x_k^{i}$ as $x_k^{i} = \{w_{kr}^i~|~ r=1,\ldots,|x_k^{i}|\}$ and similarly $\hat{x}_k^{i\rightarrow j} = \{w_{kr^{\prime}}^{{i\rightarrow j}}~|~ r^{\prime}=1,\ldots,|x_k^{i\rightarrow j}|\}$. Utilizing the attention mechanism, we can establish the correspondence $(r, r^\prime)$ between the words. Among different pairings of such words we select only the ones where $w_{kr}^i$ is a noun (e.g., as detected by a POS tagger),and enforce that the corresponding transferred word $w_{kr}^{i\rightarrow j}$ matches that noun, i.e.,
\begin{align}
\label{eq:cnt_loss}
\mathcal L_{cnt\_rec} = \mathbb{E}_{x_k^j=\{\ldots, w_{kr}^j\ldots\}\sim X} \left[-\log p_G\left(x_k^j=\{\ldots, w_{kr^j}\ldots\}| E(x_k^i, s_i), s_j\right)\right],
\end{align}
for indices $r$ and $r^\prime$ such that $w_{kr}^i$ is a noun and $(r, r^\prime)$ is a pair established by attention mechanism. We note that although not always applicable, the above heuristic is very effective for attributes where the sentences can share the nouns (e.g., for sentiment transfer considered in Section \ref{sec:experiments}).

\textbf{Classification Loss.} The loss is formulated as follows:
\begin{align}
\label{eq:class_td}
\mathcal L_{class\_td} &= \mathbb{E}_{\hat{x}_k^{i \rightarrow j}\sim \hat{X}} \left[- ~\log p_C(s_j|\hat{x}_k^{i\rightarrow j})\right].
\end{align}
For the encoder-decoder this loss gives a feedback on the current generator's effectiveness on transferring sentences to a new attribute.
For the classifier, it provides an additional training signal from generated data, enabling the classifier to be trained in a semi-supervised regime.

\textbf{Classification Loss - Original Data.} In order to enforce a high classification accuracy, the classifier also uses a supervised classification loss, measuring the classifier predictions on the original (supervised) instances $x_{k}^{i} \in X$:
\begin{align}
\label{eq:class_od}
\mathcal L_{class\_od} = \mathbb{E}_{x_k^i\sim X} \left[- ~\log p_C(s_i|x_k^i)\right].
\end{align}

\subsection{Backward Transfer}
\textbf{Reconstruction Loss.} The \emph{back-transfer (or cycle) loss} \citep{zhu17,he16} is motivated by the difficulty of imposing constraints on the transferred sentences. Back-transfer transforms the transferred sentences $\hat{x}_k^{i\rightarrow j}$ back to the original attribute $s_i$, i.e., $\hat{x}_k^{i\rightarrow j \rightarrow i}$ and compares them to $x_k^i$. This also implicitly imposes the constraints on the generated sentences and improves the content preservation (in addition to \eqref{eq:cnt_loss}). The loss is formulated as follows:
%If the content is not preserved in the transferred sentence,
%the model will not be able to reconstruct the original sentence in the back-transfer step.
%The word sequence reconstruction loss in the backward transfer case  can be formulated as follows:
\begin{align}
\label{eq:back_rec_loss}
\mathcal L_{back\_rec} = \mathbb{E}_{x_k^i\sim X} \left[- ~\log p_G(x_k^i| E(\hat{x}_k^{i\rightarrow j}, s_j), s_i)\right],
\end{align}
which can be thought to be similar to an auto-encoder loss in \eqref{eq:rec_loss} but in the attribute domain.

\textbf{Classification Loss.}
Finally, we ensure that the back-transferred sentences $\hat{x}_k^{i\rightarrow j \rightarrow i}$ have the correct attribute label $s_i$:
\begin{align}
\label{eq:class_btd}
\mathcal L_{class\_btd} &= \mathbb{E}_{\hat{x}_k^{i \rightarrow j}\sim \hat{X}} \left[- ~\log p_C(s_i|G(E(\hat{x}_k^{i\rightarrow j}, s_j),s_i))\right]
\end{align}

In summary, the training of the components of our architecture consists in optimizing the following loss function using stochastic gradient descent with back-propagation for some weights $\lambda_i>0$:
\begin{align}
\label{eq:total_loss}
&\mathcal L(\theta_E,\theta_G,\theta_{C}) = \nonumber\\
&=\min_{E, G, C} ~ \mathcal \lambda_1L_{rec} + \lambda_2L_{cnt\_rec} +\lambda_3\mathcal L_{back\_rec} + \lambda_4\mathcal L_{class\_od} + \lambda_5\mathcal L_{class\_td} + \lambda_6\mathcal L_{class\_btd}.
\end{align}
The Algorithm \ref{alg:train} summarizes the above discussion and shows the main steps of the training of the proposed approach.

\begin{algorithm}[!th]
\caption{Training of the Neural Text Attribute Transfer Algorithm using Non-parallel Data.}
\label{alg:train}
\begin{algorithmic}
\REQUIRE Two non-parallel corpora $X = X_0 \cup X_1$ with different attribute values $s_0$ and $s_1$.
\STATE Initialize $\theta_E,\theta_G,\theta_{C}$
\REPEAT

\STATE --- Sample a mini-batch of $l$ original sentences $A=\{x_k^{i}\}_{k=1}^l$ from $X$, with $i \in \{0,1\}$

\STATE --- Sample a mini-batch of $l$ transferred sentences $B=\{\hat{ x}_k^{i\rightarrow j}\}_{k=1}^l$ from the generator's distribution $p_G$, where $\hat{x}_k^{i\rightarrow j} = G(E( x_k^i,s_i),s_j)$ with $i,j \in \{0,1\}$

\STATE --- Sample a mini-batch of $l$ back-transferred sentences $C=\{\hat{ x}_k^{i\rightarrow j \rightarrow i}\}_{k=1}^l$ from the generator's distribution $p_G$, where $\hat{x}_k^{i\rightarrow j \rightarrow i} = G(E(x_k^{i\rightarrow j}, s_j),s_i)$ with $i,j \in \{0,1\}$

\STATE --- Compute $\mathcal L_{\text{rec}}$ \eqref{eq:rec_loss}, $\mathcal L_{cnt\_rec}$ \eqref{eq:cnt_loss}, $\mathcal L_{class\_td}$ \eqref{eq:class_td}, $\mathcal L_{class\_od}$ \eqref{eq:class_od}, $\mathcal L_{back\_rec}$ \eqref{eq:back_rec_loss}, and $\mathcal L_{class\_btd}$ \eqref{eq:class_btd}

\STATE --- Update $\{\theta_E,\theta_G \theta_C\}$ by gradient descent on loss $\mathcal L(\theta_E,\theta_G,\theta_{C})$ in Eq. \eqref{eq:total_loss}
\UNTIL{convergence}
\end{algorithmic}
\end{algorithm}

%\vspace{-5pt}

% \begin{figure}[!t]
% 	\centering
% 	\includegraphics[width=0.4\textwidth]{Figures/model_small.pdf}
% 	\caption{A simplified view of the proposed neural style transfer algorithm. Given sentences in style 0/1, the algorithm transfers them to style 1/0. The model is trained on non-parallel text corpus by imposing two types of losses: (i) the word reconstruction loss between the original and transferred sentences (ii) the classification loss between sentences transferred to different styles.}
% 	\label{fig:model_small}
% \end{figure}

% \begin{figure}[!t]
% 	\centering
% 	\includegraphics[width=0.95\textwidth]{Figures/model_big.pdf}
% 	\caption{A detailed view of the proposed neural style transfer algorithm.}
% 	\label{fig:model_big}
% \end{figure}

% , i.e., there are no examples of the same sentences that have both styles,

%(e.g., it can be the original $x_k^{i}$ or the transferred $\hat{x}_k^{i\rightarrow j}$ sentence)

\section{Related Work}
\label{sec:related_work}
%We must inform why our solutions to problems (a) and (b) described in the intro are different from what has been previously proposed in the literature.
%In the work of \cite{shen17} the authors proposed ..
%,taigman2016unsupervised
\vspace{-5pt}

Attribute transfer has been studied more extensively in the context of images than in the text domain, with several works studying the style transfer task under the setting of non-parallel data \citep{gatys2016image,zhu2017unpaired}. However, style/attribute transfer in text is fundamentally different as textual data is sequential and of potentially varying lengths, versus constant-sized images. In the image domain, one of the similar works is CycleGAN \citep{zhu2017unpaired}, which also employs a cycle consistency loss (similar to our \emph{back-transfer loss}) that ensures that composition of a transfer and its reverse is close to the identity map. However, there are several key differences between CycleGAN and our work: (i) we use of a single generator for generating both styles which makes it easier to scale to multiple style transfer, (ii) we use a collaborative classifier for measuring the style instead of a adversarial discriminator, which imparts stability to the training, (iii) additional syntactic regularizers for better content preservation.

%1. do the networks we use to understand language capture stylistic information in a meaningful way (note that this is largely dependent on corpus and network structure), 2. where is that style encoded, and 3. how do we transfer on sequences of changing length? Generating style-transformed sequences will likely look different than how transfer is done on images, and it's plausible we see something like a translation-style machine that takes an encoder reading in a paragraph and a decoder which is trained on a particular author that attempts to read the paragraph out in that author's style or voice.

Controlled text generation and style transfer without parallel data has also received attention from the language community recently \citep{mueller2017sequence,hu2017towards,ficler2017controlling,shen17}. \citet{ficler2017controlling} consider the problem of attribute conditioned generation of text in a conditioned language modeling setting using LSTM. \citet{mueller2017sequence} allows modifying the hidden representations to generate sentences with desired attributes which is measured by a classifier, however their model does not explicitly encourage content preservation. Our proposed model has some similarities with the approach taken by \citet{hu2017towards} and \citet{shen17}, with the main differences being that instead of VAE and adversarial discriminators we use a simple encode-decoder framework with a collaborative classifier augmented with the attention mechanism and a set of specially designed content preservation losses.

\section{Experiments and Results}
\label{sec:experiments}
%\vspace{-5pt}
In this Section we present experimental results of applying the proposed approach for sentiment transfer as one example of text attribute transfer. We compared the algorithm with the approach of \cite{shen17} on two datasets. One is the dataset from \cite{shen17}, which is based on Yelp restaurant reviews and contains (179K, 25K, 51K) sentences for (training, validation, testing) based on negative reviews and similarly (268K, 38K, 76K) positive sentences. The sentences had a maximum length of 17 words. The second dataset is based on general customer reviews on Amazon \cite{amz_data}, from which we selected (265K, 33K, 33K) positive and the same number of negative sentences, each having up to 7 tokens per sentence.

\begin{table}[!htbp]
\centering
\caption{Evaluation results on Yelp and Amazon datasets. For Yelp, the pre-trained classifier had a default accuracy of $97.4\%$ and the pre-trained language model had a default perplexity of $23.5$. For Amazon, these values were $82.02\%$ for classification and $25.5$ for perplexity.}
\label{tbl:yelp_amz}
\begin{tabular}{l|lll|lll}
\hline
& & Yelp & & & Amazon \\
\hline
 & Sentiment & Content & Perplexity & Sentiment & Content & Perplexity \\ \hline
Shen~et. al \cite{shen17}  & 86.5 & 38.3 & 27.0 & 32. 8 & 71.6 & 27.3\\
Our Method  & 94.4 & 77.1 & 80.1 & 59.5 & 77.5 & 43.7 \\ \hline
\end{tabular}
\end{table}
%\vspace{-5pt}

\begin{table}[!htbp]
\centering
\caption{Examples of sentences transferred from positive to negative sentiment on Yelp dataset}
\label{yelp}
\begin{tabular}{ll|l}
\hline
Original &  their food was definitely delicious & love the southwestern burger\\
\hline
\hline
 Shen~et. al \cite{shen17}  & there was so not spectacular &  avoid the pizza sucks  \\
Our Method  &  their food was never disgusting & avoid the grease burger\\ \hline
\hline
Original &  restaurant is romantic and quiet & the facilities are amazing\\
\hline
\hline
Shen~et. al \cite{shen17}  & the pizza is like we were disappointed &  the drinks are gone  \\
Our Method  &  restaurant is shame and unprofessional & the facilities are ridiculous\\ \hline
\end{tabular}
\end{table}
%\vspace{-5pt}

\begin{table}[!htbp]
\centering
\caption{Examples of sentences transferred from negative to positive sentiment on Yelp dataset}
\label{yelp_sent}
\begin{tabular}{ll|l}
\hline
Original&  sorry they closed so many stores & these people will try to screw you over\\
\hline
\hline
\cite{shen17}  & thanks and also are wonderful &  these guys will go to work  \\
Ours  &  amazing they had so many stores & these people will try to thank you special\\ \hline
\hline
Original&  i wish i could give them zero stars & seriously , that 's just rude\\
\hline
\hline
\cite{shen17}  & i wish i love this place &  clean , and delicious ...  \\
Ours  &  i wish i 'll give them recommended stars & seriously , that 's always friendly\\ \hline
\end{tabular}
\end{table}

We used three evaluation metrics: (i) sentiment accuracy, which is computed based on pre-trained classifier (estimated on the training part of each dataset) and measures the percentage of sentences in the test set with correct sentiment label; (ii) content preservation accuracy, a new evaluation metric proposed in this work, which is computed as the percentage of the transferred sentences where each of them has at least one of the nouns present in the original sentence; (iii) perplexity score, which is computed based on pre-trained language model (estimated on the training part of each dataset) and measures the quality of the generated text. 

The results are presented in Table \ref{tbl:yelp_amz}. As compared to the algorithm of Shen et. al \cite{shen17}, the proposed method although not able to get better perplexity scores, it can achieve more accurate sentiment transfer and better content preservation. A possible explanation of having higher perplexity is that since the algorithm of \cite{shen17} does not explicitly enforce content similarity, it has an easier job of achieving high sentiment accuracy and low perplexity of the transferred sentences. Our algorithm, on the other hand, is penalized if the content changes, which forces it to sacrifice the perplexity. Achieving better results across all the metrics still remains a challenge.

%Observe that an algorithm that can output the transferred sentence as a copy of the original sentence can achieve the lowest possible perplexity and $100\%$ of content accuracy - this approach would, however, have a very low sentiment accuracy.   

In Table \ref{yelp_sent} we also show some of the sentences generated by both algorithms on Yelp dataset. The algorithm of \cite{shen17}, although able to create well structured sentences with correct sentiment labels, in many cases it cannot accurately preserve the content. On the other hand, our approach may generate text with somewhat higher perplexity but ensures a better sentiment and content transfer.

%\subsubsection*{Acknowledgments}
\section{Conclusion}
In this work we proposed a novel algorithm for text attribute transfer with non-parallel corpora based on the encoder-decoder architecture with attention, augmented with the collaborative classifier and a set of content preservation losses. Although the experimental evaluations showed promising results, a number of challenges remain: (i) achieve better results across all the three metrics and propose new evaluation metrics to better capture the quality of transfer; (ii) improve the architecture to enable transfer for more challenging text attributes (e.g., such as professional-colloquial) where the text goes under more significant transformation then in a simpler sentiment transfer tasks; (iii) extend the architecture to work in a multi-attribute transfer, a more challenging problem.

\bibliographystyle{abbrvnat} % or try plainnat or unsrtnat
\bibliography{references} % refers to example.bib

%\appendix
%\input{appendix}

\end{document}